\def\year{2021}\relax
\documentclass[letterpaper]{article}
\usepackage{aaai21} % DO NOT CHANGE THIS
\usepackage{times} % DO NOT CHANGE THIS
\usepackage{helvet} % DO NOT CHANGE THIS
\usepackage{courier} % DO NOT CHANGE THIS
\usepackage[hyphens]{url} % DO NOT CHANGE THIS
\usepackage{graphicx} % DO NOT CHANGE THIS
\usepackage{graphicx}  % DO NOT CHANGE THIS
\usepackage{natbib}
\usepackage{caption}
\usepackage{amsmath}
\usepackage{amsfonts}
\usepackage[ruled, vlined, linesnumbered]{algorithm2e}
\usepackage{subcaption}
\title{Rule-based Shielding for Partially Observable Monte-Carlo Planning}
\author{
    Giulio Mazzi,
    Alberto Castellini,
    Alessandro Farinelli\\
}
\affiliations{
    Università degli studi di Verona,\\
    Dipartimento di Informatica,\\
    Strada Le Grazie 15, 37134, Verona, Italy\\
    giulio.mazzi@univr.it, alberto.castellini@univr.it, alessandro.farinelli@univr.it
}

\DeclareRobustCommand{\freevar}[1]{\overline{\textbf{#1}}}

\begin{document}
\maketitle 

\begin{abstract}
Partially Observable Monte-Carlo Planning (POMCP) is a powerful online algorithm able to generate approximate policies for large Partially Observable Markov Decision Processes. The online nature of this method supports scalability by avoiding complete policy representation. The lack of an explicit representation however hinders policy interpretability and makes policy verification very complex. In this work, we propose two contributions. The first is a method for identifying unexpected actions selected by POMCP with respect to expert prior knowledge of the task. The second is a shielding approach that prevents POMCP from selecting unexpected actions. The first method is based on Satisfiability Modulo Theory (SMT). It inspects traces (i.e., sequences of belief-action-observation triplets) generated by POMCP to compute the parameters of logical formulas about policy properties defined by the expert. The second contribution is a module that uses online the logical formulas to identify anomalous actions selected by POMCP and substitutes those actions with actions that satisfy the logical formulas fulfilling expert knowledge. We evaluate our approach on Tiger, a standard benchmark for POMDPs, and a real-world problem related to velocity regulation in mobile robot navigation. Results show that the shielded POMCP outperforms the standard POMCP in a case study in which a wrong parameter of POMCP makes it select wrong actions from time to time. Moreover, we show that the approach keeps good performance also if the parameters of the logical formula are optimized using trajectories containing some wrong actions.
\end{abstract}

\section{Introduction} \label{sec:introduction}
Planning in partially observable environments while satisfying safety guarantees is a challenging problem.
\emph{Partially Observable Markov Decision Processes (POMDPs)}~\cite{Cassandra97} is a popular framework to model systems with uncertainty.
 Computing an optimal solution for POMDPs is very hard (i.e., PSPACE-complete~\cite{Papadimitriou1987}). However, it is possible to compute an approximate solution, and state-of-the-art algorithms achieve great performance in real-world instances of POMDPs.
A pioneering algorithm for this purpose is \emph{Partially Observable Monte-Carlo Planning (POMCP)}~\cite{Silver2010} which uses a particle filter to represent the belief and a Monte-Carlo Tree Search based strategy to compute the policy online.
The online nature of the policy, however, makes the task of analyzing the decisions taken by POMCP very difficult~\cite{Castellini2020,Castellini2019AIRO,Castellini2019}.
In general, with a high number of particles POMCP yields great performance, but sometimes the simulation does not properly assess the risk of certain actions, especially if the number of particles used in the simulation is limited due to engineering constraints (e.g., time limits, agents with limited computation capabilities or the model of the real environment is approximated).
Moreover, in POMCP the policy is never fully computed or stored, hence it is very difficult to identify the reasons for possible unexpected decisions of the system.
Explainability~\cite{Gunning2019} is becoming a key feature of artificial intelligence systems since in these contexts humans need to understand why specific decisions are taken by the agent.
Specifically, explainable planning (XAIP)~\cite{Fox2017, Cashmore2019} focuses on explainability in planning methods.
The presence of erroneous behaviors in these tools (due, for instance, to the wrong setup of internal parameters) may have a strong impact on autonomous cyber-physical and robotic systems that interact with humans, and detecting these errors in automatically generated policies is very hard in practice.

In this work, we propose a methodology for generating a safety mechanism from high-level descriptions of the desired behavior of a POMCP-generated policy.
In this approach, a human expert provides qualitative information on a property of the system, enriched with an indication of the expected behavior that the system should have in specific situations (e.g., ``the robot should move fast if it is highly confident that the path is not cluttered, I expect this level of confidence to be above 90\%'').
With this information, our methodology analyzes a set of execution traces of the system and provides quantitative details of these statements by analyzing the execution of the system (e.g., ``the robot moves fast if its confidence of being in an uncluttered segment is at least $93.4\%$).
The approach we propose formalizes the problem of parameters computation as a \emph{MAX-SMT} problem which allows to express complex logical formulas and to compute optimal assignments when the template is not fully satisfiable (which happens in the majority of cases in real policy analysis).
This quantitative answer is then used to build a shield, namely a safety mechanism that forces the POMCP to satisfy the constraints expressed by the expert.
The shield works alongside the Monte-Carlo Tree Search, by preemptively blocking some actions that, according to the rules, should not be selected in some situations.
The shield is also enriched with a mechanism that quantifies how much the current belief is far from satisfying the rule.
This allows some extra flexibility in filtering the beliefs that are acceptable by the shield, an important requirement for real-time algorithms that can only work with partial knowledge.

To empirically evaluate the performance of the shield, we test it in two domains, namely, the well-known \emph{Tiger} problem and a robotic problem in which a mobile platform must move as fast as possible in a cluttered environment avoiding collisions.
To test the robustness of the shield mechanism, we inject an error into POMCP by wrongly setting one of its parameters.
This error subtly affects the decisions of the policy keeping almost unchanged the average performance but making some decisions non-optimal and therefore unexpected by experts (i.e., unsafe).
Our experiments show that the shielding mechanism can improve performance while using a compact and expressive representation of expected properties. We also show that the use of the shield degrades performance only negligibly in practice.

In summary, the contribution of this paper to the state-of-the-art is threefold:
\begin{itemize}
    \item we propose an SMT-based methodology that combines a logic-based description of a system with the real execution traces of a POMCP policy to create a set of rules describing expected behaviors of an agent;
    \item we propose a method for generating a shield from this set of rules to block unexpected actions;
    \item we empirically evaluate the shielding mechanism in two domains showing that it can exploit the knowledge provided by the expert to achieve higher performance than standard POMCP when its parameters are imprecise.
    %rewards w.r.t. the unshielded POMCP method.
\end{itemize}
The paper is structured as follows.
In section \ref{sec:related_works} we present related work.
In section~\ref{sec:background} provide background knowledge. 
In section~\ref{sec:method} we describe our methodology, and in Section~\ref{sec:results} we show the performance of our method.
Finally, in Section~\ref{sec:conclusion} we summarize our contribution and we provide direction for future development.

\section{Related Work} \label{sec:related_works}
This work is mainly related to two other fields in the literature, namely, the verification of POMDPs policies and explainable planning. A possible approach for policy verification consists of encoding the POMDP problem into a standard logic-based framework, such as those presented in~\cite{Cashmore2016, Norman2017, Wang2018, Bastani2018}, and then prove property guarantees formally using an SMT-solver. However, these frameworks present scalability problems due to the computational complexity of solving large instances of SMT. In our work, we add a shielding mechanism on top of POMCP preventing performance degradation in large problems. This is reached by combining the logic-based representation of the problem with the highly efficient online approach provided by POMCP. 

In~\cite{Zhu2019} verification is also achieved by exploiting a simplified representation of the problem provided by experts.
The method verifies properties related to the safety of fully observable systems modeled by Markov Decision Processes (MDPs).
It then works on a pre-trained neural network representing a black-box policy and uses a linear formula summarizing the policy behavior to use off-the-shelf verification tools.
This differs from our approach because we work directly on partially observable environments, hence we explicitly consider beliefs (instead of states) in our logic-based rules describing the expected behavior of the policy.
Moreover, our methodology is able to work also on subsets of actions, and it does not require rules describing the dynamic of each action.

Another approach for verifying POMCP properties is presented in~\cite{Newaz2019}, it uses \emph{statistical model checking (SMC)} to verify that qualitative objectives, specified on possible states of the environment, are satisfied with a certain confidence level.
This is different from our methodology since we specify property on beliefs instead of on states. 
SMC requires also a large number of particles to generate reliable results while our methodology does not have such a requirement.

The need for a shielding mechanism based on expert expectation for highly complex planning approaches is highlighted by the growing interest in the field of \emph{eXplainable Artificial Intelligence (XAI)}~\cite{Gunning2019}, a rapidly growing research field focusing on human interpretability and understanding of artificial intelligence (AI) systems.
In particular, our work is related to explainable planning (XAIP)~\cite{Cashmore2019, Fox2017, Langley17, Sule2019} that aims at developing planning tools whose decisions are understandable by human beings.
In our work, we use the high-level insight provided by the user as a guide for building rules that summarize the expected policy behavior.
A particularly interesting kind of questions analyzed in XAIP is known as \emph{contrastive questions}~\cite{Fox2017}.
An example is ``Why have you made decision $d_1$ instead of $d_2$ , that I expect to be a better option?''.
An explainable system should be able to provide a human-comprehensible answer motivating its choice and providing human-understandable evidence that it is better than the alternative option.
These questions are, however, very difficult to answer in online frameworks, as POMCP, because the information required to build the answer may not be available to the agent at run time ~\cite{Castellini2020}.
For instance, the fact that POMCP computes the policy only for beliefs encountered in its execution, and it does not generate an explicit representation of the overall policy (as other methods do with $\alpha$-vectors for example) prevent the usage of methods that analyze this explicit representation of the policy to explain it.
Our method does not use contrastive questions but it bases the interaction between human and planner on logical formulas that are local representations of the policy constrained by the expert's prior knowledge and specialized by the SMT solver on previously observed behaviors of the planner (contained in traces). 
A methodology focused on building explainable rules that describe as many of the decisions taken by the policy as possible is presented in~\cite{Mazzi2021AAMAS}.
It identifies the decisions that do not satisfy the rule (i.e., unexpected decisions) to improve the explainability of the results.
This differs from our work because it is an offline procedure that cannot be used to improve the performance of a POMCP algorithm.
The integration of a rule-based explanation into a shield allows our method to identify in real-time decisions that violate user's expectations and avoiding them improves the planner performance.

\section{Background} \label{sec:background}
In this section, we present a definition of the POMDP framework and the POMCP algorithm.
We also briefly present the SMT problem and the MAX-SMT extension.

\subsection{Partially Observable Monte-Carlo Planning}
A Partially Observable Markov Decision Process (POMDP)~\cite{Kaelbling98} is a tuple $(S, A, O, T, Z, R, \gamma)$,
where $S$ is a set of partially observable \emph{states},
$A$ is a set of \emph{actions},
$Z$ is a finite set of \emph{observations},
$T$:~$S\times A \rightarrow \Pi(S)$ is the \textit{state-transition model}, with $\Pi(S)$ probability distribution over states,
$O$:~$S\times A \rightarrow \Pi(Z)$ is the \textit{observation model},
$R$:~$S\times A \rightarrow \mathbb{R}$ is the \textit{reward function} and $\gamma \in [0,1]$ is a \textit{discount factor}.
An agent must maximize the \emph{discounted return} $E[\sum_{t=0}^{\infty} \gamma^t R(s_t,a_t)]$.
A probability distribution over states, called \emph{belief}, is used to represent the partial observability of the true state.
To solve a POMDP it is required to find a \emph{policy}, namely a function $\pi$:~$B \rightarrow A$ that maps beliefs $B$ into actions.

In this work, we focus on \emph{Partially Observable Monte-Carlo Planning (POMCP)}~\cite{Silver2010} to solve POMDPs.
POMCP is an \emph{online} algorithm that solves POMDPs by using Monte-Carlo techniques.
The strength of POMCP is that it does not require an explicit definition of the transition model, observation model, and reward.
Instead, it uses a black-box to simulate the environment.
POMCP uses a \emph{Monte-Carlo Tree Search} (MCTS) at each time-step to explore the belief space and select the best action.
\emph{Upper Confidence Bound for Trees (UCT)}~\cite{Kocsis2006} is used as a search strategy to select the subtrees to explore and balance exploration and exploitation.
The belief is implemented as a \emph{particle filter}, which is a sampling over the possible states that is updated at every step.
At each time-step, a particle (which represents a specific state of the POMDP) is selected from the filter.
This corresponding state is used as an initial point to perform a simulation in the Monte-Carlo tree.
Each simulation is a sequence of action-observation pairs and it collects a discounted return, and for each action, we can compute the expected reward that can be achieved.
After receiving an observation, the particle filter is updated to reflect the new information available in the environment.
If required, new particles can be generated from the current state through a process of \emph{particle reinvigoration}.
In the following, we call \emph{trace} a set of runs performed by POMCP on a specific problem. Each run is a sequence of \emph{steps}, and each step corresponds to an action performed by the agent having a belief and receiving an observation from the environment.

\subsection{SMT and MAX-SMT}
The problem of reasoning on the satisfiability of formulas involving propositional logic and first-order theories is called  \emph{Satisfiability Modulo Theory} (SMT).
In our methodology, we use propositional logic and the theory of linear real arithmetic to encode the information provided by the expert, and we use Z3~\cite{DeMoura2008} as an SMT-solver to compute the parameter of the shield that codify these ideas.
Specifically, we encode our formulas as a MAX-SMT problem, which has two kinds of clauses, namely, \emph{hard}, that must be satisfied, and \emph{soft} that should be satisfied whenever possible.
A model of the MAX-SMT problem hence satisfies all the hard clauses and as many soft clauses as possible, and it is unsatisfiable only when hard clauses are unsatisfiable.
In our methodology, the expert provides a high-level explanation (that can include abstractions and local approximation) and it is intended to describe as many decisions as possible among those taken by the policy, hence MAX-SMT provides a perfect formalism to encode this requirement.
The Z3 solver is used to solve the MAX-SMT problem~\cite{Bjorner2014}. Subsection~\ref{subsec:shieldgen} presents the details of this encoding.

\section{Methods} \label{sec:method}

\begin{figure}[!t]
\centering
\includegraphics[width=1.0\columnwidth]{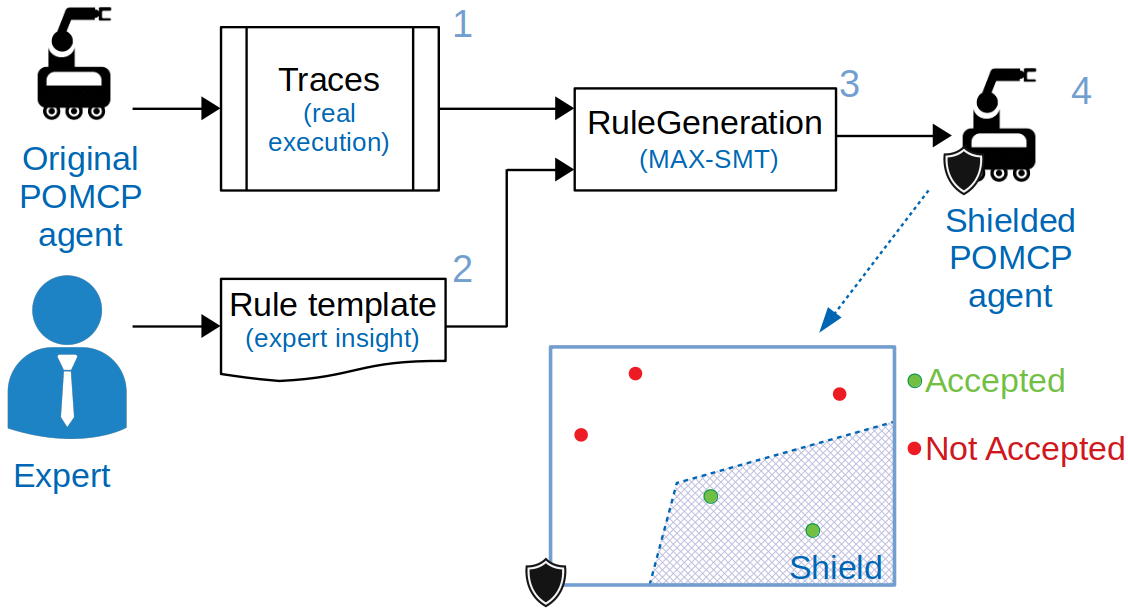} 
\caption{Methodology overview.} \label{fig:method_overview}
\end{figure}
The methodology proposed in this work is summarized in Figure~\ref{fig:method_overview}.
It leverages the expressiveness of logical formulas to represent specific properties of the system under investigation, and this representation is used to automatically generate a \emph{shield}, a security mechanism that forces the POMCP system to satisfy a set of high-level requirement.

As a first step, a logical formula with free variables is defined (see box $2$ in Figure~\ref{fig:method_overview}) to describe a property of interest of the policy under investigation. This formula, called \emph{rule template}, defines a relationship between some properties of the belief (e.g., the probability to be in a specific state) and an action.
Free variables in the formula allow the expert to avoid quantifying the limits of this relationship. These limits are then computed by analyzing a set of observed traces (see box 1). For instance, a template saying ``Do this when the confidence that the path is cluttered is at least $\freevar{x}$'', with $\freevar{x}$ free variable, is transformed into ``Do this when the confidence that the path is cluttered is at least $0.15$''.
By defining a rule template the expert provides useful prior knowledge about the structure of the investigated property.
This template can include strict requirements that the expert wants to satisfy (e.g., it is possible to force the rule to only use confidence values above $90\%$).
This is combined with the real execution of a POMCP system collected into a trace.
The methodology trains a \emph{rule} (i.e., a rule template with all the free-variables instantiated) using a MAX-SMT based algorithm. This rule describes as many of the decisions taken by the agent as possible while satisfying the requirement defined in the template (see box 3).
A set of rules is then used to create a \emph{shield}, a safety mechanism that we integrate into POMCP to preemptively block actions that do not respect the details defined by the expert with the template (box 4).

\subsection{Rules and Rule Templates}
A rule template encodes the structure of the expected behavior of the system.
It can include hard requirements that the expert expects to be true and free-variables that the methodology must instantiate.
It is a set of first-order logic formulas without quantifiers explaining some properties of the policy, and has the following form:
\begin{align*}
    \begin{split}
        & \texttt{$r_1:$ select $a_1$ when (${\bigvee}_{i^1}$ subformula$_{i^1}$);} \\
        & \dots\\
        & \texttt{$r_n:$ select $a_n$ when (${\bigvee}_{i^n}$ subformula$_{i^n}$);} \\
        & \texttt{{[ where~${\bigwedge}_j$ (requirement$_j$); ]}}
    \end{split}
\end{align*}
where $r_1,\dots,r_n$ are \emph{action rule templates}.
A \emph{subformula} is defined as ${\bigwedge}_k p_{s} \approx \freevar{x}_k$, where $p_s$ is the probability of state $s$, symbol $\approx \in \{<, >, \ge, \le\}$, and $\freevar{x}_k$ is a free variable that is automatically instantiated by the SMT solver analyzing the traces (when the problem is satisfiable).
In general, bold letters with an overline (e.g., $\freevar{x}, \freevar{y}$) are used to identify free variable while italic letters (e.g., $p, a_i$) are used for fixed values read from the trace.
The \texttt{where} statement is used to specify hard requirements. They can take different forms, such a the definition of a minimum value (e.g., $\freevar{x}_0 \ge 0.9$) or a relation (e.g., $\freevar{x}_2 = \freevar{x}_3$).
These are used to define prior knowledge on the domain which is used by the rule generation algorithm to compute optimal parameter values (e.g., equality between two free-variables belonging to different rules can be used to encode the idea that two rules are symmetrical).

\subsection{Rule Generation} \label{subsec:shieldgen}
The rules used in the shield are generated using Algorithm~\ref{algo:rule_generation}, it takes as input a trace \emph{ex} generated by the POMCP system that we want to shield and a rule template \emph{r}. The output is an instantiation of all the free-variables of \emph{r} that satisfies as many steps of the step of \emph{ex} as possible.
\begin{algorithm}[ht]
    \KwData{a trace generated by POMCP $ex$\\\hspace{0.965cm}a rule template $r$}
    \KwResult{an instantiation of $r$}
    $solver \leftarrow$ probability constraints for thresholds in $r$\; \label{line:probability_axioms}
    \ForEach{action rule $r_a$ with $a\in A$} {\label{line:begin_dummy}
        \ForEach{step $t$ in $ex$} {
            build new dummy literal $l_{a,t}$\; \label{line:init_literal}
            $cost \leftarrow cost \cup l_{a,t}$\; \label{line:init_cost}
            compute $p_0^t, \dots, p_n^t$ from \emph{t.particles}\; \label{line:particles}
            $r_{a,t} \leftarrow $ instantiate rule $r_a$ using $p_0^t, \dots, p_n^t$\;\label{line:instatiation}
            \If{$t.action \neq a$} { 
                $r_{a,t} \leftarrow \lnot(r_{a,t})$\;\label{line:negation}
            }
            $solver.add(l_{a,t} \lor r_{a,t})$\;
             \label{line:end_if_action}
        } 
    } \label{line:end_dummy}
    
    $solver$.minimize($cost$)\; \label{line:minimize_cost}
    $goodness \leftarrow 1 - distance\_to\_observed\_boundary$\; \label{line:interval}
    $model \leftarrow$ $solver$.maximize($goodness$)\; \label{line:maximize_rule}
    \Return{$model$} \label{line:return_rule}
    \caption{RuleGeneration} \label{algo:rule_generation}
\end{algorithm}
A Z3 instance (\emph{solver}) is initialized in line~\ref{line:probability_axioms}.
Hard constraints are added to force all the free-variables in the template to satisfy the probability axioms (i.e., to have value in range $[0, 1]$).
Then in the \emph{foreach} loop in lines \ref{line:begin_dummy}--\ref{line:end_dummy} the algorithm builds a solution that satisfy the maximum number of steps in trace $ex$ using the template $r$.
In particular, for each action rule $r_a$, where $a$ is an action, and for each step $t$ in the trace $ex$, the algorithm first generates a literal $l_{a,t}$ (line \ref{line:init_literal}) which is a dummy variable used by MAX-SMT to satisfy clauses that are not satisfiable by a free variable assignment.
This literal is then added to the \emph{cost} objective function (line \ref{line:init_cost}) which is a pseudo-boolean function collecting all the dummy literals.
Afterwards, the belief state probabilities are collected from the particle filter (line~\ref{line:particles}) and used to instantiate the action rule template $r_a$ (line \ref{line:instatiation}) by substituting their probability variables $p_i$ with observed belief probabilities.
This generates a new clause $r_{a,t}$ which represents the constraint for step $t$.
This constraint is considered in its negated form $\lnot(r_{a,t})$ if the step action $t.action$ is different from $a$ (line \ref{line:negation}) because the clause $r_{a,t}$ should not be true.
To encode the fact that we need the maximum number of satisfiable steps, we add the clause $l_{a,t} \lor r_{a,t}$ to the problem. 
These clauses can be satisfied in two ways, namely, by finding an assignment of the free variables that makes the clause $r_{a,t}$ true or by assigning a true value to the literal $l_{a,t}$.
To minimize the number of occurance of the second case we use the solver to find a solution that minimize the value of the cost function.
This minimization is a typical MAX-SMT problem in which an assignment maximizing the number of satisfied clauses is found. 
There can be more than a single assignment of free variables that achieves the MAX-SMT goal, thus the last step of the algorithm (lines~\ref{line:interval}--\ref{line:maximize_rule}) concerns the identification of the assignment which is closer to the behavior observed in the trace.
This problem is solved by maximizing a goodness function which moves the free variables assignment as close as possible to the numbers observed in the trace, without altering the truth assignment of the dummy literals.
Notice that this problem concerns the maximization of a real function, not the maximization of the number of satisfiable clauses (as in MAX-SMT). It is solved by the linear arithmetic module of the Z3 solver.

The variable in the SMT problem are the free variables specified in the template (a constant number) and the dummy literals, that are linear on the size of the trace because the algorithm builds a clause for each step, and each clause introduces a new dummy literal.
MAX-SMT is NP-hard but in practice, Z3 solves most practical instances in a reasonable time, and it provides good performance in our experiments (For example, it never takes more than one minute to compute the parameter of a rule in the more challenging case of velocity regulation).

\subsection{Soft-thresholding for Rule Membership Check} \label{sec:rule_points}

It is important to retain flexibility in the shield mechanism for two reasons, namely, to allow the user to express approximate ideas (important for explainability) and to be capable of considering unexplored beliefs (important because POMCP is an online algorithm).
The rule describes as many of the steps of the trace as possible, however, the MAX-SMT-based solution does not explain, in general, all the decisions taken by the policy due to the approximate nature of the template used to generate the rule.
Some of these decisions are very different from the template (i.e., very different than the behavior expected by the expert), and thus should not be accepted by the shield, but others may be only slightly outside the rule boundaries (possibly because the approximation does not work well in this step).
For example, if we have the rule ``move at high speed when the probability of collision is below $5\%$'', the shield must not accept a decision to move at high speed when the risk of collision is $40\%$, but a belief in which we only have a $5.2\%$ risk of collision could be considered (e.g., because other aspects that were omitted in the simplified template play a role in this decision).
Moreover, the online nature of POMCP does not allow us to write a rule that describes the whole belief space due to the incomplete nature of the policy generated by this method.
Training the rule using a large trace that contains many runs of the original POMCP helps in reducing this problem, but it is important to maintain flexibility toward unexplored beliefs.

To foster flexibility, we introduce a mechanism that quantifies how far a belief is from the rule, and if this difference is lower than a threshold $\tau$ we accept the belief as valid.
To compute the \emph{distance} between a new belief and the rule, we choose some beliefs that satisfy the rule as representatives of the rule itself, and we measure the \emph{discrete Hellinger distance ($H^2$)} \cite{Hellinger1909} between the new belief (whose membership to the rule is under evaluation) and each representative.
Given two discrete probability distributions $P$ and $Q$ over $k$ states (in our case, the two beliefs under consideration) the Hellinger distance is defined as:
\begin{gather*}
    H^2(P, Q) = \frac{1}{\sqrt{2}} \sqrt{\sum_{i = 1}^{k} ( \sqrt{p_i} - \sqrt{q_i} )^2}
\end{gather*}
where $p_i$ is the probability of the $i$-th state in $P$ and $q_i$ is the probability used to generate the rule of the $i$-th state in $Q$.
An interesting property of $H^2$ is that it is bounded between $0$ and $1$, which is very useful to define a meaningful threshold $\tau$.
The rule describes one or more constraints on the acceptable beliefs, we use these constraints to generate a set $D$ of $d$ possible beliefs that satisfy the rule.
In other words, when the rule membership of a new belief $b$ is checked by the shield, if the belief is out of the rule decision boundary then we compute the $H^2$ between $b$ and each belief in $D$. Then we compare all the distances with the threshold $\tau$.
If one of the distances is below the threshold then the shield considers $b$ as an acceptable belief for the rule.

\subsection{Shielded POMCP} \label{sec:shield_decisions}

\begin{algorithm}[t]
    \KwData{a belief $b$, a shield $s$, safe action $a_{safe}$}
    \KwResult{set of legal actions $\mathcal{L}$}
    $\mathcal{L} \leftarrow \emptyset$\;
    \ForEach{action $a\in \mathcal{A}$} {
        \If{$a \notin s$} {\label{line:norule}
            $\mathcal{L} \leftarrow \mathcal{L} \cup a$\;
        }
        \ElseIf{$s.\texttt{test\_constraints}(b)$} { \label{line:constraints}
            $\mathcal{L} \leftarrow \mathcal{L} \cup a$\;
        }
        \ElseIf{$\exists r \in s.Repr : H^2(b, r) < s.\tau$} {\label{line:representative}
            $\mathcal{L} \leftarrow \mathcal{L} \cup a$\;
        }
    }
    \If{$\mathcal{L} = \emptyset$} {
        $\mathcal{L} \leftarrow \{a_{safe}\}$\;
    }
    \Return{$\mathcal{L}$} \label{line:return_legal_actions}\;
    \caption{Shielding} \label{algo:shielding}
\end{algorithm}

We integrate the shield into POMCP to preemptively prune undesired actions considering the current belief.
It includes a set of rules trained as explained in Section~\ref{subsec:shieldgen} and a set of representative beliefs generated as described in Section~\ref{sec:rule_points}.
To shield the action of the POMCP, we start by building a set of \emph{legal actions} $\mathcal{L}$ that satisfy the logical rules and can be performed on the current belief, and we force POMCP to only consider legal actions in the first step of the simulation.
After a legal action is selected, the Monte-Carlo Tree Search is performed as usual.
Notice that when the original implementation of POMCP~\cite{Silver2010} selects a particle in the simulation process, it assumes that the state encoded by the particle is the current state of the system (which for a POMDP is not observable) and thus the belief can only be considered in the first step.

With this mechanism, we can ensure that the rule of the shield is respected but we do not force POMCP to select a specific action, the best action is still decided using the regular POMCP but only among the legal ones.
In more detail, as reported in Algorithm~\ref{algo:shielding}, for each possible action $a$, we consider $a$ as a legal action if it satisfies at least one of these three conditions, namely, \emph{i)} the shield does not define any rule (i.e., any restriction) for this action (line~\ref{line:norule}), \emph{ii)} the current belief satisfies the constraints defined by an action rule for action $a$ (line~\ref{line:constraints}), \emph{iii)} the Hellinger distance between the belief and the closest representative of the action rule for $a$ is lower than the predefined threshold (line~\ref{line:representative}).
These conditions could result in an empty set of legal actions $\mathcal{L}$ (i.e., if the rules are very strict).
In this case, it is important to define a default safe action $a_{safe}$ that is used when no other action is possible.
While this is a domain-specific requirement, it is reasonable to assume that most domains have such action (e.g., wait and listen to gather extra data, take low-risk action that yields low rewards).
The computation of legal actions is performed only once for a simulation step since the current belief does not change until a new observation is received from the real environment.
Checking that the belief satisfies the constraints has a fixed cost, checking the $H^2$ of the representative beliefs increases linearly with the number of beliefs.
As shown in Section~\ref{sec:results}, this is a negligible cost, and the reduced number of actions that must be tested (because not all actions are now legal) can also slightly reduce the execution time.

\section{Results} \label{sec:results}
\subsection{Case Studies} \label{subsec:case_studies}
We test our methodology in two domains, namely, the standard POMDP domain \emph{tiger} and a robotic-inspired problem called \emph{velocity regulation} in the following, in which a robot must move as fast as possible while avoiding collisions.

\paragraph{Tiger} \emph{Tiger} is a well-known problem~\cite{Kaelbling98} in which an agent has to chose which door to open among two doors, one hiding a treasure and the other hiding a tiger. Finding the treasure yields a reward of $+10$ while finding the tiger a reward of $-100$. The agent can also listen (by paying a small penalty of $-1$) to gain new information. Listening is however not accurate since there is a $0.15$ probability of hearing the tiger from the wrong door.

\begin{figure}[t]
\centering
\includegraphics[width=1.0\columnwidth]{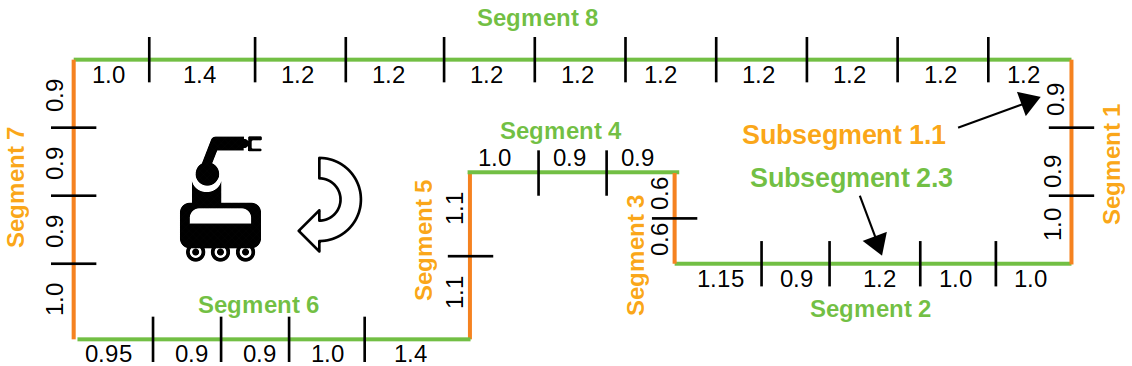} 
\textbf{(a)}
\begin{minipage}{.50\linewidth}
\centering
\small
\begin{tabular}{ccc}
    \hline
    $f$ & \hspace{0.2cm} & $p(o=1 \ | \ f)$ \\
    \hline
    0 & & 0.44\\
    1 & & 0.79\\
    2 & & 0.86\\
    \hline
     & & \textbf{(b)\phantom{111}}\\
\end{tabular}
\end{minipage}%
\begin{minipage}{.50\linewidth}
\small
\centering
\begin{tabular}{rrr}
    \hline
    $f$ & $a$  & $p(c=1 \ | \ f,a)$ \\
    \hline
    0 & 0 & 0.0\\
    0 & 1 & 0.0\\
    0 & 2 & 0.028\\
    1 & 0 & 0.0\\
    1 & 1 & 0.056\\
    1 & 2 & 0.11\\
    2 & 0 & 0.0\\
    2 & 1 & 0.14\\
    2 & 2 & 0.25\\
    \hline
    & & \textbf{(c)\phantom{111111}}\\
    \end{tabular}
\end{minipage}
\caption{Main elements of the POMDP model for \emph{velocity regulation}. (a) Path map. The map presents the length (in meters) for each subsegment. (b) Occupancy model $p(o \ | \ f)$: probability of observing a subsegment occupancy given segment difficulty. (c) Collision model $p(c\ | \ f,a)$: collision probability given segment difficulty and action.} \label{fig:lab_ice}
\end{figure}

\paragraph{Velocity Regulation} In \emph{velocity regulation}, a robot travels on a pre-specified path divided into eight \emph{segments} which are in turn divided into \emph{subsegments} of different sizes, as shown in Figure~\ref{fig:lab_ice}.
Each segment has a (hidden) difficulty value among \emph{clear} ($f = 0$, where $f$ is used to identify the difficulty), \emph{lightly obstructed} ($f = 1$) or \emph{heavily obstructed} ($f = 2$). 
All the subsegments in a segment share the same difficulty value, hence the hidden state-space has $3^8$ states.
The goal of the robot is to travel on this path as fast as possible while avoiding collisions.
In each subsegment, the robot must decide a \emph{speed level} $a$ (i.e., action).
We consider three different speed levels, namely $0$ (slow), $1$ (medium speed), and $2$ (fast).
The reward received for traversing a subsegment is equal to the length of the subsegment multiplied by $1+a$, where $a$ is the speed of the agent, namely the action that it selects.
The higher the speed, the higher the reward, but a higher speed suffers a greater risk of collision (see the collision probability table $p(c=1 \ | \ f,a)$ in Figure \ref{fig:lab_ice}.c).
The real difficulty of each segment is unknown to the robot, but in each subsegment, the robot receives an observation, which is $0$ (no obstacles) or $1$ (obstacles) with a probability depending on segment difficulty (see Figure \ref{fig:lab_ice}.b). 
The state of the problem contains a hidden variable (i.e., the difficulty of each segment), and three observable variables (current segment, subsegment, and time elapsed since the beginning).

\subsection{Empirical Methodology}
We implement \emph{tiger} and \emph{velocity regulation} as black-box simulators in the original POMCP C++ code provided in~\cite{Silver2010}.
We extend the implementation with the capability of using a shield.
To generate \emph{traces}, we collect (belief, action) pairs at each step, with the belief saved as a particle distribution.
We store these data using the \emph{eXtensible Event Stream (XES)}~\cite{XES2017} format, a standard developed to log the executions of programs conveniently.
The \emph{RuleGeneration} algorithm (i.e., Algorithm~\ref{algo:rule_generation}) has been developed in Python.
The Python binding of Z3~\cite{DeMoura2008} has been used to solve the SMT formulas.
To build the shields, we use rules that prove to be good explanations of the policies under investigation.
We build these rules using the explainability tool presented in~\cite{Mazzi2021AAMAS}. A detailed example of the rule synthesis procedure is presented in~\cite{Mazzi2020AIRO}.
Experiments have been performed on a notebook with Intel Core i7-6700HQ and 16GB RAM.
The code is available at \emph{https://github.com/GiuMaz/ICAPS-2021-supmat}.

\paragraph{Error Injection}
To test the robustness of the shield in different scenarios, we injected an error in the POMCP implementation of the two domains.
We modify the \emph{RewardRange} parameter (called $c$ in the following) in POMCP.
This parameter defines the maximum difference between the lowest and the highest possible reward, and it is used by UCT to balance exploration and exploitation.
If this value is lower than the correct one the algorithm could find a reward that exceeds the maximum expected value leading to a wrong state, namely, the agent believes to have identified the best possible action and it stops exploring new actions, even though the selected action is not the best one.
This is an interesting error because it is hard to detect, it randomly affects the exploration-exploitation trade-off without introducing any systematic (and thus, easier to identify) mistakes.
This erroneous behavior is also independent of the domain used, and thus it is useful to recreate similar situations in different problems.
Moreover, notice that the error we consider can happen with a relevant frequency in practical applications of POMCP, as the parameter $c$ is not easy to evaluate automatically without exhaustively considering all the possible states of the POMDP, and thus must be tuned by the designer.

In our experimental setup, we consider several possible values of the parameter $c$ and for each domain and value of $c$ we create a trace using the unshielded POMCP, then we train a shield for each generated execution trace by using the methodology described in Section~\ref{sec:method}.
Notice that these traces could contain errors.
Finally, we run the POMCP again using the shielding and the various values of \emph{c}.
We evaluate the performance of the methodology by comparing the average discounted return achieved by the two versions of POMCP (original and shielded). We consider $1000$ runs for \emph{Tiger} and $100$ runs for \emph{Velocity Regulation}.
The results achieved for tiger and velocity regulation are presented in subsections~\ref{subsec:tiger_experiments} and \ref{subsec:vr_experiments}, respectively.
In both cases, we generate 1000 representative beliefs and we use $\tau=0.10$ as a threshold to check if a belief is in range.

\subsection{Results for Tiger} \label{subsec:tiger_experiments}

\begin{table*}[ht] 
\small
\centering

\begin{tabular}{r|rr|rrrr}
\multicolumn{1}{l|}{} & \multicolumn{2}{c|}{No Shield} & \multicolumn{4}{c}{Shield} \\ \hline
\multicolumn{1}{c|}{\emph{c}} & \multicolumn{1}{c}{return} & \multicolumn{1}{c|}{time (s)} & \multicolumn{1}{c}{return} & \multicolumn{1}{c}{RI} & \multicolumn{1}{c}{time (s)} &\multicolumn{1}{c}{\#SA} \\ \hline
110 & $3.702 (\pm 0.623)$  & $0.066 (\pm 0.027)$ & $3.702 (\pm 0.623)$ & $0.00\%$   & $0.065 (\pm 0.029)$  & $0$   \\
80  & $3.593 (\pm 0.632)$  & $0.067 (\pm 0.030)$ & \textbf{3.702 ($\pm$ 0.623)} & $3.03\%$   & $0.061 (\pm 0.027)$  & $4$   \\
60  & $3.088 (\pm 0.673)$  & $0.060 (\pm 0.025)$ & \textbf{3.702 ($\pm$ 0.623)} & $19.88\%$  & $0.061 (\pm 0.027)$  & $121$ \\
40  & $-4.173 (\pm 1.101)$ & $0.035 (\pm 0.017)$ & \textbf{3.702 ($\pm$ 0.623)} & $188.71\%$ & $0.052 (\pm 0.023)$  & $647$ \\
\end{tabular}

\emph{a)} Tiger

\begin{tabular}{r|rr|rrrr}
\multicolumn{1}{l|}{} & \multicolumn{2}{c|}{No Shield} & \multicolumn{4}{c}{Shield} \\ \hline
\multicolumn{1}{c|}{\emph{c}} & \multicolumn{1}{c}{return} & \multicolumn{1}{c|}{time (s)} & \multicolumn{1}{c}{return} & \multicolumn{1}{c}{RI} & \multicolumn{1}{c}{time (s)} &\multicolumn{1}{c}{\#SA} \\ \hline
103 & $24.716 (\pm 3.497)$ & $10.166 (\pm 0.682)$ & \textbf{26.045 ($\pm$ 3.640)} & $5.38\%$   & $10.118 (\pm 0.238)$ & $7$   \\
90  & $18.030 (\pm 3.794)$ & $10.173 (\pm 0.234)$ & \textbf{22.680 ($\pm$ 3.524)} & $25.79\%$  & $10.166 (\pm 0.241)$ & $12$  \\
70  & $4.943 (\pm 5.260)$  & $10.278 (\pm 0.234)$ & \textbf{8.970 ($\pm$ 4.556)}  & $81.46\%$  & $10.377 (\pm 0.230)$ & $51$  \\
50  & $0.692 (\pm 5.051)$  & $10.374 (\pm 0.230)$ & $1.638 (\pm 4.525)$  & $136.53\%$ & $10.435 (\pm 0.336)$  & $171$ \\
\end{tabular}

\emph{b)} Velocity Regulation

\caption{Experimental Results. The first column shows the different values of the RewardRange \emph{c}. The second (third) column shows the average return (time) achieved by the original POMCP and the relative standard deviation. The \emph{Shield} section shows the average return and time achieved by POMCP using a shield (column four and six), values in bold show a statistically significant difference with respect to the shield counterpart (according to a paired t-test with 95\% confidence level). Column \emph{RI} shows the relative increase in performance between the two original and shielded POMCP. Finally, column \emph{\#SA} shows how many times the shield alters the decision during the execution} \label{tab:experimental_results}
\end{table*}

A successful policy for \emph{Tiger} listens to collect information on the position of the tiger and opens a door only when the agent is reasonably certain to find the treasure.
From the analysis of the observation model and reward function, it is however not immediate to define what ``reasonably certain'' means.
To investigate it, we create a rule template specifying a relationship between the confidence (in the belief) over the treasure position and the related opening action as follow:
\begin{align*}
    \begin{split}
    & \texttt{$r_L:$ select}~Listen~\texttt{when~} (p_{right} \le \freevar{x}_1 \land p_{left} \le \freevar{x}_2);\\
    & \texttt{$r_{OR}:$ select}~Open_{R}~\texttt{when~} p_{right} \ge \freevar{x}_3;\\
    & \texttt{$r_{OL}:$ select}~Open_{L}~\texttt{when~} p_{left} \ge \freevar{x}_4;\\
    & \texttt{where } (\freevar{x}_1 = \freevar{x}_2) \land (\freevar{x}_3 = \freevar{x}_4) \land (\freevar{x}_3 > 0.9);
    \end{split}
\end{align*}
Action rule template $r_L$ describes when the agent should listen, while templates $r_{OR}$ and $r_{OL}$ describe when the agent should open the right and left door, respectively.
We also want to force the agent to only open a door if it is at least 90\% sure to find the treasure behind it ($\freevar{x}_3 > 0.9$).
Finally, we use two hard constraints to specify that the problem is expected to be symmetric ($x_1 = x_2$ and $x_3 = x_4$).

We learn the rule parameters from a POMCP trace and we create a shield from this rule.
Since this shield gives a rule for all possible actions, it is important to set a safe action as described in Section~\ref{sec:shield_decisions}. For this domain, we use $a_{safe} = Listen$.
The correct value of \emph{c} is $110$ because the reward interval is $[-100,10]$.
For each value of $c$ in $\{110, 80, 60, 40\}$, we generate a trace with $1000$ runs each, using a fixed seed for the pseudo-random algorithm.
In each case, we use $2^{15}$ particles and a maximum of $10$ steps.
POMCP with a correct value of \emph{c} produces the optimal policy (we tested that by comparing the decisions taken by POMCP with an exact policy computed using \emph{incremental pruning}~\cite{Cassandra97}).
As shown in Table~\ref{tab:experimental_results}.a, in particular in the first row of column \emph{\#SA} (the meaning of each column name is defined in the table caption), the shield does not interfere with the correct policy.
Lower values of $c$ produce a lower average discounted return, as shown in column \emph{return} of the \emph{No Shield} section in Table~\ref{tab:experimental_results}.a.
In both sections \emph{No Shield} and \emph{Shield} of Table~\ref{tab:experimental_results}.a, we present the average \emph{return} and the average execution \emph{time} of POMCP.
In the \emph{Shield} section we also present the relative increase (i.e., $RI = \frac{shielded - original}{|original|}\cdot 100$) between the original and the shielded version of the POMCP (see column \emph{RI}).
This column shows that the benefit of using a shield is greater when the number of errors increases.
For the return column, we report in bold values whose difference from their no-shield counterpart is statistically significant according to a paired t-test with 95\% confidence.
While in the first row there is no difference between the two cases, in the other three rows the difference is statistically significant.

The average return achieved using the shield is the same in all four cases, and this is also identical to the return achieved by the correct policy. This is because in \emph{tiger} we can write a shield that perfectly recreates the behavior of the correct policy, a goal that is difficult to achieve in real-world problems.
This is particularly interesting because the shields in the cases of $c \in \{80, 60, 40\}$ are obtained by using traces generated with a POMCP implementation that does make some mistakes. As a consequence, the Execution traces contain wrong decisions. However, the combination of insight provided by the expert with the MAX-SMT-based analysis of the traces results in a shield with extremely good performances.
As shown in the \emph{time} column, in general, the presence of the shield does not noticeably impact in terms of run-time.
The shield generation algorithm takes between 10 and 12 seconds to generate the shield in this case.
In the last row, the original POMCP is particularly fast, this happens since the erroneous POMCP opens many doors as fast as possible, without listening, this leads to runs that achieve very low average return but ends quickly.

\subsection{Results for Velocity Regulation} \label{subsec:vr_experiments}

In \emph{velocity regulation}, a robot moves around a predefined path in an industrial environment.
As in \emph{Tiger}, we use $2^{15}$ particles.
In general, this leads to acceptable performance but sometimes the simulations are not good enough and the robot takes a decision that the designer considers too risky.
We focus on writing a shield describing when the robot travels at maximum speed (i.e., $a=2$).
This is the most dangerous action because there is always a risk of collision involved, as explained in Table~\ref{fig:lab_ice}.c.
We expect that the robot should move at speed $2$ only if it is confident enough to be in an easy-to-navigate segment, but this level of confidence varies slightly from segment to segment (due to the length of the segments, the elapsed times, or the relative difficulty of the current segment in comparison to the others).
To write a rule template that is compact but informative, we want the rule to be a local approximation of the behavior of the robot, thus we only focus on the current segment without considering the path as a whole.
To do that, we introduce the \texttt{diff} function, which, given a belief, a segment, and a difficulty value, extracts the expected difficulty of the segment in the specified belief.
We can not write a compact template:
\begin{align*}
    & \texttt{$r_2:$ select action $S_{2}$ when~} \\
    & \hspace{0.5cm}  p_0 \ge \freevar{x}_1 \lor p_2 \le \freevar{x}_2 \lor (p_0 \ge \freevar{x}_3 \land p_1 \ge \freevar{x}_4)\\
    & \texttt{where}~ p_i = \texttt{diff(distr, seg, i)}\forall i \in \{0,1,2\} \\
    & \hspace{1.2cm} \land \freevar{x}_1 \ge 0.9
\end{align*}
where $\freevar{x}_1,\freevar{x}_2,\freevar{x}_3,\freevar{x}_4$ are free variables.
With the subformula $p_0 \ge \freevar{x}_1$ we say that the robot can move at high speed if its confidence of being in an easy to navigate (i.e., difficulty 0) segment is above a certain threshold. In the hard requirement, we also force this threshold to be at least $0.9$.
The robot can also move at high speed if it is confident of not being in a really hard segment (subformula $\freevar{x}_2$), or if the combination of $p_0$ and $p_1$ (i.e., clear or slightly cluttered segment) are above other thresholds (subformula $p_0 \ge \freevar{x}_3 \land p_1 \ge \freevar{x}_4$).
We do not force any requirement for these other thresholds, we simply train the values from the data.
We use a trace with $100$ runs to train and test the shield.
The shield generation takes $~50$ seconds to generate velocity regulation shields.
Table~\ref{tab:experimental_results}.b shows the result of the experiment.
As in \emph{Tiger}, a lower value of \emph{c} produce a lower return.
In this case, the best \emph{c} is $103$ (the difference between a collision when we move slowly in the shortest segment and a fast movement in the longest segment without a collision).
The first row shows that, unlike \emph{Tiger}, the usage of a shield can improve the performance even when \emph{c} is correct.
In this case, the shield intervenes only $7$ times (over the $3500$ analyzed steps), yielding a $5.38\%$ increment in the return.
This happens because the shield blocks the rare cases in which the POMCP simulations are not enough to properly assess the risk of moving at high speed.
When \emph{c} decreases, the shield intervenes more often (see column \emph{\#SA}) since the error due to the limited number of simulations is combined with the errors generated by an incorrect value of \emph{c}.
Table~\ref{tab:experimental_results}.b also shows that a higher number of interventions leads to a bigger relative increase in the performance (column \emph{RI}).
The difference is statistically significant in the case of $c \in \{103, 90, 70\}$, and show that the introduction of the shield improves the performance up to the $81\%$, even in cases in which the shield is trained using traces generated by a POMCP process that makes some mistakes.
In the case of $c=50$ the return increase but the difference is not statistically significant.
The shield intervenes $171$ times by blocking risky high-speed moves, but unlike \emph{Tiger}, in which we use a rule for every possible action, here POMCP made many wrong decisions when it moves at low or medium speed (for example, by moving slowly when the path is clear).
The usage of the shield does not significantly increase the time required to perform the simulations.

\section{Conclusions and Future Work} \label{sec:conclusion}
In this work, we present a methodology that generates a shielding mechanism for POMCP exploiting a high-level representation of expected policy behavior provided by human experts. The shielding mechanism preemptively blocks unexpected actions.
The approach is proved to provide a statistically significant improvement to the performance of POMCP in a standard domain and a robotics-inspired domain.
This work paves the way towards several interesting research directions.
First, we aim at improving the expressiveness of logical formulas by employing temporal logic to verify safe reachability requirements.
Second, we aim to further improve the integration between POMCP and the shielding mechanism (e.g., by considering the effect of shielding on other actions besides the first one of the simulation).
Third, most important, we aim at developing an approach for synthesizing logical rules online, i.e., while the POMCP algorithm is running and not based only on previously generated traces.

%\bigskip
%
%\bigskip
 
% References and End of Paper
% These lines must be placed at the end of your paper
\bibliography{main} \label{sec:bib}

\begin{thebibliography}{25}
\providecommand{\natexlab}[1]{#1}
\providecommand{\url}[1]{\texttt{#1}}
\providecommand{\urlprefix}{URL }
\expandafter\ifx\csname urlstyle\endcsname\relax
  \providecommand{\doi}[1]{doi:\discretionary{}{}{}#1}\else
  \providecommand{\doi}{doi:\discretionary{}{}{}\begingroup
  \urlstyle{rm}\Url}\fi

\bibitem[{{Acampora} et~al.(2017){Acampora}, {Vitiello}, {Di Stefano}, {van der
  Aalst}, {Gunther}, and {Verbeek}}]{XES2017}
{Acampora}, G.; {Vitiello}, A.; {Di Stefano}, B.; {van der Aalst}, W.;
  {Gunther}, C.; and {Verbeek}, E. 2017.
\newblock IEEE 1849: The XES Standard: The Second IEEE Standard Sponsored by
  IEEE Computational Intelligence Society [Society Briefs].
\newblock \emph{IEEE Computational Intelligence Magazine} .

\bibitem[{Anjomshoae et~al.(2019)Anjomshoae, Najjar, Calvaresi, and
  Fr\"{a}mling}]{Sule2019}
Anjomshoae, S.; Najjar, A.; Calvaresi, D.; and Fr\"{a}mling, K. 2019.
\newblock Explainable {A}gents and {R}obots: {R}esults from a {S}ystematic
  {L}iterature {R}eview.
\newblock In \emph{Proceedings of the 18th International Conference on
  Autonomous Agents and MultiAgent Systems}, AAMAS '19. Richland, SC.

\bibitem[{Bastani, Pu, and Solar-Lezama(2018)}]{Bastani2018}
Bastani, O.; Pu, Y.; and Solar-Lezama, A. 2018.
\newblock Verifiable {R}einforcement {L}earning via {P}olicy {E}xtraction.
\newblock In \emph{Proceedings of the 32nd International Conference on Neural
  Information Processing Systems}, NIPS’18, 2499–2509. Red Hook, NY, USA.

\bibitem[{Bj\o{}rner, Phan, and Fleckenstein(2015)}]{Bjorner2014}
Bj\o{}rner, N.; Phan, A.-D.; and Fleckenstein, L. 2015.
\newblock v{Z} - {A}n {O}ptimizing {SMT} {S}olver.
\newblock In \emph{Proceedings of the 21st International Conference on Tools
  and Algorithms for the Construction and Analysis of Systems - Volume 9035},
  194–199. Berlin, Heidelberg.

\bibitem[{Cashmore et~al.(2019)Cashmore, Collins, Krarup, Krivic, Magazzeni,
  and Smith}]{Cashmore2019}
Cashmore, M.; Collins, A.; Krarup, B.; Krivic, S.; Magazzeni, D.; and Smith, D.
  2019.
\newblock Towards {E}xplainable {AI} {P}lanning as a {S}ervice.
\newblock 2nd ICAPS Workshop on Explainable Planning, XAIP 2019.

\bibitem[{Cashmore et~al.(2016)Cashmore, Fox, Long, and
  Magazzeni}]{Cashmore2016}
Cashmore, M.; Fox, M.; Long, D.; and Magazzeni, D. 2016.
\newblock {A} {C}ompilation of the {F}ull {PDDL}+ {L}anguage into {SMT}.
\newblock In \emph{Proceedings of the {T}wenty-{S}ixth {I}nternational
  {C}onference on {I}nternational {C}onference on {A}utomated {P}lanning and
  {S}cheduling}, {ICAPS}'16, 79--87.

\bibitem[{Cassandra, Littman, and Zhang(1997)}]{Cassandra97}
Cassandra, A.; Littman, M.~L.; and Zhang, N.~L. 1997.
\newblock Incremental {P}runing: {A} {S}imple, {F}ast, {E}xact {M}ethod for
  {P}artially {O}bservable {M}arkov {D}ecision {P}rocesses.
\newblock In \emph{In Proceedings of the Thirteenth Conference on Uncertainty
  in Artificial Intelligence}, 54--61.

\bibitem[{{C}astellini, {C}halkiadakis, and {F}arinelli(2019)}]{Castellini2019}
{C}astellini, A.; {C}halkiadakis, G.; and {F}arinelli, A. 2019.
\newblock {I}nfluence of {S}tate-{V}ariable {C}onstraints on {P}artially
  {O}bservable {M}onte {C}arlo {P}lanning.
\newblock In \emph{{P}roc. 28-th {I}nternational {J}oint {C}onference on
  {A}rtificial {I}ntelligence, {IJCAI-19}}, 5540--5546.

\bibitem[{Castellini, Marchesini, and Farinelli(2020)}]{Castellini2019AIRO}
Castellini, A.; Marchesini, E.; and Farinelli, A. 2020.
\newblock Online Monte Carlo Planning for Autonomous Robots: Exploiting Prior
  Knowledge on Task Similarities.
\newblock In \emph{Proceedings of the 6th Italian Workshop on Artificial
  Intelligence and Robotics (AIRO 2019@AI*IA2019)}, volume 2594 of \emph{{CEUR}
  Workshop Proceedings}, 25--32. CEUR-WS.org.

\bibitem[{Castellini et~al.(2020)Castellini, Marchesini, Mazzi, and
  Farinelli}]{Castellini2020}
Castellini, A.; Marchesini, E.; Mazzi, G.; and Farinelli, A. 2020.
\newblock Explaining the influence of prior knowledge on {POMCP} policies.
\newblock In \emph{Proceedings of the 17th European Conference on Multi-Agents
  Systems}, volume 12520 of \emph{Lecture Notes in Artificial Intelligence}.

\bibitem[{De~Moura and Bj\o{}rner(2008)}]{DeMoura2008}
De~Moura, L.; and Bj\o{}rner, N. 2008.
\newblock Z3: {A}n {E}fficient {SMT} {S}olver.
\newblock In \emph{Proceedings of the Theory and Practice of Software, 14th
  International Conference on Tools and Algorithms for the Construction and
  Analysis of Systems}, TACAS’08/ETAPS’08, 337–340. Berlin, Heidelberg.

\bibitem[{Fox, Long, and Magazzeni(2017)}]{Fox2017}
Fox, M.; Long, D.; and Magazzeni, D. 2017.
\newblock Explainable {P}lanning.
\newblock \emph{CoRR} abs/1709.10256.

\bibitem[{Gunning(2019)}]{Gunning2019}
Gunning, D. 2019.
\newblock {DARPA}’s {E}xplainable {A}rtificial {I}ntelligence ({XAI})
  {P}rogram.
\newblock ii--ii.

\bibitem[{Hellinger(1909)}]{Hellinger1909}
Hellinger, E. 1909.
\newblock Neue begr{\"u}ndung der theorie quadratischer formen von
  unendlichvielen ver{\"a}nderlichen.
\newblock \emph{Journal f{\"u}r die reine und angewandte Mathematik} 136:
  210--271.

\bibitem[{Kaelbling, Littman, and Cassandra(1998)}]{Kaelbling98}
Kaelbling, L.~P.; Littman, M.~L.; and Cassandra, A.~R. 1998.
\newblock Planning and {A}cting in {P}artially {O}bservable {S}tochastic
  {D}omains.
\newblock \emph{Artif. Intell.} 101(1–2): 99–134.

\bibitem[{Kocsis and Szepesv\'{a}ri(2006)}]{Kocsis2006}
Kocsis, L.; and Szepesv\'{a}ri, C. 2006.
\newblock Bandit {B}ased {M}onte-{C}arlo {P}lanning.
\newblock In \emph{Proc. ECML'06}, 282--293. Berlin, Heidelberg.

\bibitem[{Langley et~al.(2017)Langley, Meadows, Sridharan, and
  Choi}]{Langley17}
Langley, P.; Meadows, B.; Sridharan, M.; and Choi, D. 2017.
\newblock Explainable {A}gency for {I}ntelligent {A}utonomous {S}ystems.
\newblock In \emph{Proceedings of the Thirty-First AAAI Conference on
  Artificial Intelligence}, AAAI'17, 4762–4763.

\bibitem[{Mazzi, Castellini, and
  Farinelli(2021{\natexlab{a}})}]{Mazzi2021AAMAS}
Mazzi, G.; Castellini, A.; and Farinelli, A. 2021{\natexlab{a}}.
\newblock Identification of {U}nexpected {D}ecisions in {P}artially
  {O}bservable {M}onte {C}arlo {P}lanning: {A} {R}ule-{B}ased {A}pproach.
\newblock In \emph{accepted at the 21th International Conference on Autonomous
  Agents and MultiAgent Systems}, AAMAS '21.

\bibitem[{Mazzi, Castellini, and Farinelli(2021{\natexlab{b}})}]{Mazzi2020AIRO}
Mazzi, G.; Castellini, A.; and Farinelli, A. 2021{\natexlab{b}}.
\newblock Policy Interpretation for Partially Observable Monte-Carlo Planning:
  a Rule-based Approach.
\newblock In \emph{Proceedings of the 7th Italian Workshop on Artificial
  Intelligence and Robotics (AIRO 2020@AI*IA2020)}, volume 2806 of \emph{{CEUR}
  Workshop Proceedings}, 44--48. CEUR-WS.org.

\bibitem[{Newaz, Chaudhuri, and Kavraki(2019)}]{Newaz2019}
Newaz, A. A.~R.; Chaudhuri, S.; and Kavraki, L.~E. 2019.
\newblock Monte-{C}arlo {P}olicy {S}ynthesis in {POMDP}s with {Q}uantitative
  and {Q}ualitative {O}bjectives.
\newblock In \emph{RSS 2019}.

\bibitem[{Norman, Parker, and Zou(2017)}]{Norman2017}
Norman, G.; Parker, D.; and Zou, X. 2017.
\newblock Verification and control of partially observable probabilistic
  systems.
\newblock \emph{Real-Time Systems} 53(3): 354--402.

\bibitem[{Papadimitriou and Tsitsiklis(1987)}]{Papadimitriou1987}
Papadimitriou, C.~H.; and Tsitsiklis, J.~N. 1987.
\newblock The {C}omplexity of {M}arkov {D}ecision {P}rocesses.
\newblock \emph{Math. Oper. Res.} 12(3): 441--450.

\bibitem[{{S}ilver and {V}eness(2010)}]{Silver2010}
{S}ilver, D.; and {V}eness, J. 2010.
\newblock Monte-{C}arlo {P}lanning in Large {POMDP}s.
\newblock In {L}afferty, J.~D.; {W}illiams, C. K.~I.; {S}hawe {T}aylor, J.;
  {Z}emel, R.~S.; and {C}ulotta, A., eds., \emph{{A}dvances in {N}eural
  {I}nformation {P}rocessing {S}ystems 23}, 2164--2172. Curran Associates, Inc.

\bibitem[{Wang, Chaudhuri, and Kavraki(2018)}]{Wang2018}
Wang, Y.; Chaudhuri, S.; and Kavraki, L.~E. 2018.
\newblock Bounded {P}olicy {S}ynthesis for {POMDP}s with {S}afe-{R}eachability
  {O}bjectives.
\newblock \emph{ArXiv} abs/1801.09780.

\bibitem[{Zhu et~al.(2019)Zhu, Xiong, Magill, and Jagannathan}]{Zhu2019}
Zhu, H.; Xiong, Z.; Magill, S.; and Jagannathan, S. 2019.
\newblock An {I}nductive {S}ynthesis {F}ramework for {V}erifiable
  {R}einforcement {L}earning.
\newblock In \emph{Proceedings of the 40th ACM SIGPLAN Conference on
  Programming Language Design and Implementation}, PLDI 2019, 686–701. New
  York, NY, USA.

\end{thebibliography}
\end{document}

\maketitle 

\section{XPOMCP implementation}
In the \texttt{code} folder of the supplementary material we provide an implementation of the shielding methodology,
a single trace that can be used to test the shielding (\texttt{dataset\_tiger\_60.xes}),
and an already generated shield that can be used to test the POMCP with the shield (\texttt{tiger\_shield\_60.txt}).
The subfolders contain a \texttt{README.md} file that explains how to use them.